\documentclass[runningheads]{ECCV/llncs}
\usepackage{graphicx}

\usepackage{tikz}
\usepackage{comment}
\usepackage{amsmath,amssymb}
\usepackage{color}

\usepackage[width=122mm,left=12mm,paperwidth=146mm,height=193mm,top=12mm,paperheight=217mm]{geometry}

\usepackage{graphicx}
\usepackage{epsfig}
\usepackage{color}
\usepackage{amsmath,mathtools}
\usepackage{amssymb}
\usepackage[htt]{hyphenat}
\usepackage{enumerate}
\usepackage{subfigure}
\usepackage{algpseudocode}
\usepackage{booktabs}
\usepackage{tabularx}
\usepackage{array}
\usepackage{bm}
\usepackage{url}
\usepackage{bbm}

\usepackage{cleveref}
\usepackage{xcolor}
\usepackage{stackengine}
\usepackage{wrapfig}
\usepackage{multirow}
\usepackage{multicol}
\usepackage{tablefootnote}
\usepackage{comment}
\usepackage{makecell}
\usepackage{subfig}
\usepackage{xspace}
\newcommand*{\eg}{e.g.\@\xspace}
\newcommand*{\ie}{i.e.\@\xspace}

\newcommand{\R}{\mathbb{R}}

\usepackage{cleveref}
\usepackage{braket}

\Crefname{assumption}{\textbf{H}\hspace{-3pt}}{\textbf{H}\hspace{-3pt}}
\crefname{assumption}{\textbf{H}}{\textbf{H}}

\Crefname{assumption}{\textbf{H}\hspace{-3pt}}{\textbf{H}\hspace{-3pt}}
\crefname{algorithm}{\text{Alg.}}{\text{Alg.}}
\crefname{assumption}{\textbf{H}}{\textbf{H}}
\crefname{equation}{\text{Eq}}{\text{Eq}}
\crefname{definition}{\text{Dfn.}}{\text{Dfn.}}
\crefname{lemma}{\text{Lemma}}{\text{Lemma}}
\crefname{dfn}{\text{Dfn.}}{\text{Dfn.}}
\crefname{thm}{\text{Thm.}}{\text{Thm.}}
\crefname{tab}{\text{Tab.}}{\text{Tab.}}
\crefname{fig}{\text{Fig.}}{\text{Fig.}}
\crefname{table}{\text{Tab.}}{\text{Tab.}}
\crefname{figure}{\text{Fig.}}{\text{Fig.}}
\crefname{section}{\text{Sec.}}{\text{Sec.}}

\begin{document}

\pagestyle{headings}
\mainmatter

\title{6D Camera Relocalization in Visually Ambiguous Extreme Environments} 
\author{Yang Zheng\inst{1} \,\,\,
Tolga Birdal\inst{2,3} \,\,\,
Fei Xia\inst{3} \,\,\,
Yanchao Yang\inst{3} \,\,\,
\\Yueqi Duan\inst{1,3} \,\,\,
Leonidas J. Guibas\inst{3} \,\,\,
}
\authorrunning{Y. Zheng et al.}
\institute{\,\inst{1} Tsinghua University \,\,\,\,\,\, \inst{2} Imperial College London  \,\,\,\,\,\, \inst{3} Stanford University}

\maketitle

\begin{abstract}
We propose a novel method to reliably estimate the pose of a camera given a sequence of images acquired in extreme environments such as deep seas or extraterrestrial terrains. Data acquired under these challenging conditions are corrupted by textureless surfaces, image degradation, and presence of repetitive and highly ambiguous structures.
When naively deployed, the state-of-the-art methods can fail in those scenarios as confirmed by our empirical analysis. In this paper, we attempt to make camera relocalization work in these extreme situations. To this end, we propose: (i) a hierarchical localization system, where we leverage temporal information and (ii) a novel environment-aware image enhancement method to boost the robustness and accuracy. Our extensive experimental results demonstrate superior performance in favor of our method under two extreme settings: localizing an autonomous underwater vehicle and localizing a planetary rover in a Mars-like desert. In addition, our method achieves comparable performance with state-of-the-art methods on the indoor benchmark (7-Scenes dataset) using only 20\% training data.
\keywords{camera relocalization, camera pose, extreme environments}
\end{abstract}

\section{Introduction}\label{sec:intro}\vspace{-3mm}
Camera relocalization considers inferring the 6DoF camera translation and orientation parameters with respect to a known 3D scene. Relocalization provides a key to open many doors and thus, it has become an essential component in various applications, \eg, autonomous navigation, augmented reality, reconstruction and etc. Various kinds of sensory input such as IMU, GPS, or LIDAR, can be used to localize an autonomous vehicle. Out of all these, visual cues coming from RGB cameras carry a special place paving the way to a general purpose, robust and accurate localization ability.

\begin{figure}[ht]
    \vspace{5pt}
    \subfigure[Common scenes]{\includegraphics[width=0.28\linewidth]{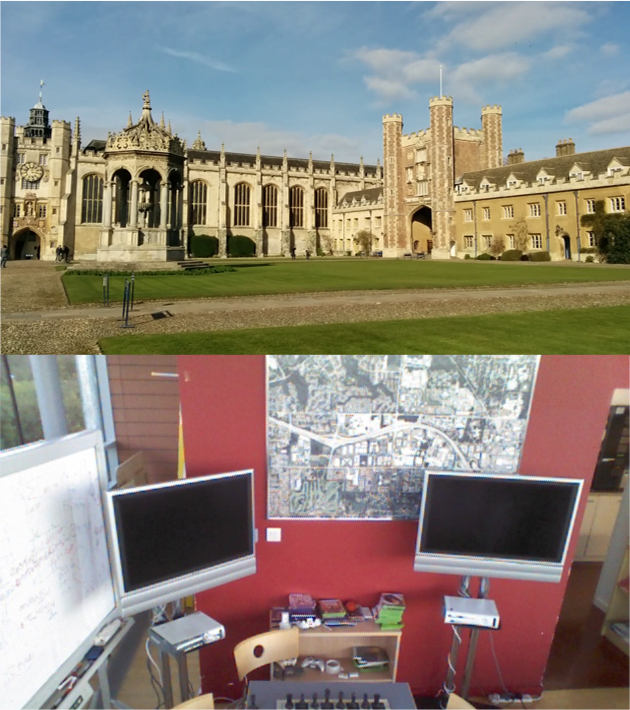}}
    \subfigure[Visually ambiguous extreme environments]{\includegraphics[width=0.72\linewidth]{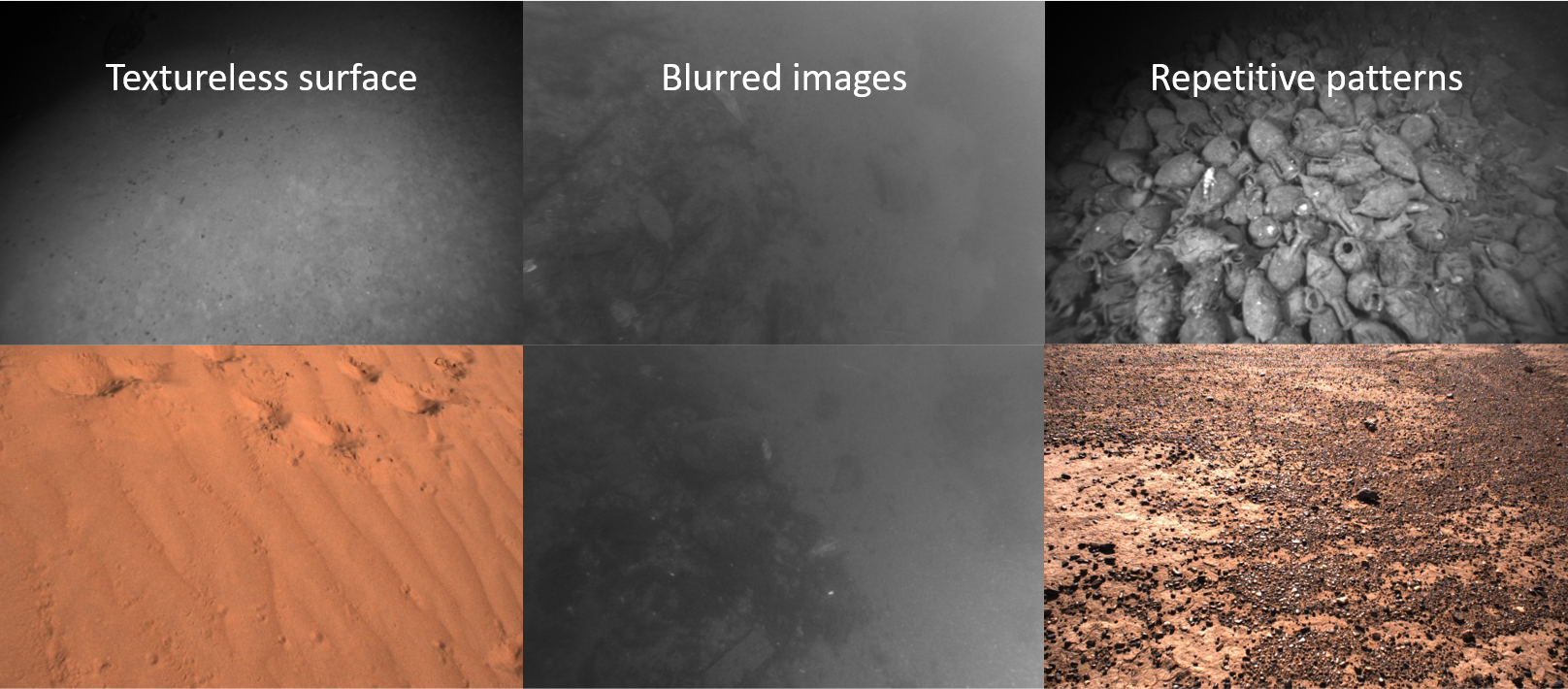}}

    \caption{\textbf{(a)} Common scenes from Cambridge landmarks~\cite{kendall2015posenet} and 7-Scenes~\cite{shotton2013scene} datasets,  
    where discriminate features can be extracted.
    \textbf{(b)} Challenging situations from underwater Aqualoc dataset~\cite{ferrera2019aqualoc} and Mars-Analogue dataset~\cite{meyer2021madmax}.
    Data are severely corrupted with ambiguous elements and low image quality. We'll show that even the state-of-the-art method designed for localization in ambiguous scenes~\cite{bui20206d}, \cite{deng2020deep} can be confused in such scenarios.\vspace{-3mm}}
    \label{fig:teaser}
    \vspace{-10pt}
\end{figure}
With the advances in modern machine learning, direct regression of a 6D camera pose from a given query image has become a popular approach for relocalization~\cite{kendall2015posenet}, ~\cite{walch2017image}, ~\cite{kendall2017geometric}, ~\cite{brahmbhatt2018geometry}, ~\cite{sattler2019understanding}. 
As a trade-off for simplicity, this type of methods can not guarantee the robustness of inference.
Thus, another branch utilizes geometric correspondences for improving the accuracy of the estimation. Once 2D/3D-3D correspondences are established by a neural decision tree~\cite{meng2017backtracking}, ~\cite{meng2018exploiting}, ~\cite{cavallari2019real}, ~\cite{dong2021robust} or retrieval-based methods~\cite{sarlin2019coarse}, the 6D camera pose can be calculated~\cite{kabsch1976solution} and optimized by RANSAC~\cite{fischler1981random}. The premise of these methods are well validated:
on the indoor benchmark 7-Scenes~\cite{shotton2013scene} (see~\cref{fig:teaser}(a)), recently introduced methods~\cite{cavallari2019real}, ~\cite{brachmann2021visual}, ~\cite{dong2021robust} can achieve nearly 100\% accuracy (poses within $5\mathrm{cm}/5^{\circ}$ error) using auxiliary information like depth maps or pre-scanned 3D models.

Unfortunately, the huge success of these state-of-the-art approaches can generalize only to indoor scenes or common outdoor environments of the cities. Oftentimes, autonomous vehicles are faced with additional challenges when they are used in environments which are not easily penetrable by humans \eg in underwater or extraterrestrial medium. Unlike our everyday environments, images acquired in these settings lack sufficient semantic and geometric information, suffer from low visibility and lighting variations, may contain ambiguous structures, and additional environmental challenges which make it difficult to deploy typical hardware such as depth sensors, GPS or laser scanners. All these impair the myriad of state-of-the-art methods in finding the cues indicating the camera pose.

In this paper, we aim to handle the challenge of camera relocalization in such visually ambiguous extreme environments which suffer from degraded image quality. As opposed to the multi-hypotheses prediction networks~\cite{bui20206d,deng2020deep} or Bayesian methods~\cite{kendall2016modelling}, which explicitly model the ambiguities, our goal is to regress the confident 6D pose robustly. To this end, under the assumption that the query images are recorded as a continuous sequence, we propose to leverage temporal information in an end-to-end, correspondence driven, deep architecture to keep the localization on track. 
The temporal adjacent frames provide a strong reference to estimate the current position, mitigating the challenges caused by ambiguous elements.
Specially, our system involves a hierarchical localization framework, which contains two coarse-to-fine localization steps. The first step is an iterative temporal matching to localize at map level. The second step is a pose refinement step. Besides robustly localizing the camera under ambiguities, our method can recover the 3D scene in a test-time SfM. 

Aside from the ambiguity, another factor that may hinder robust localization is the image quality. 
In extreme conditions, images can be degenerated due to different medium or bad illumination, which will undermine feature detection and tracking when constructing the 3D scene. 
To tackle the challenge, we propose an environment-aware image enhancement method by training a light-weight neural network through self-supervision, where the network learns to recover the latent clear image by minimizing the keypoint matching assignment loss of covisible image pairs. Experimental results clearly demonstrate the effectiveness of our design in improving localization accuracy.

In general, the main contributions in this work include:
\begin{itemize}
    \item To our best knowledge, we are the first to perform camera relocalization experiments in visually ambiguous extreme environments. We benchmark the state-of-the-art methods, and shed light on the challenges faced by current vision localization systems through extensive experiments.
    \item We propose a robust localization framework which makes advantage of temporal information to handle the localization ambiguity in such extreme environments, and leverage a pose refinement method to boost the accuracy while refining the 3D structure on the whole sequence.
    \item We design an environment-aware image enhancement module to optimally improve the image quality such that the downstream feature-based reconstruction and localization losses are minimized.
\end{itemize}

\section{Related Work}\label{sec:related} \vspace{-3mm}
\noindent\textbf{Relocalization by direct regression.} This type of method predicts the 6Dof camera pose directly from the input image.
A simple strategy is to find an approximate pose by using image retrieval~\cite{lim2012real}, ~\cite{weyand2016planet}, which depends on dataset discretization and may fail in challenging cases, \eg textureless regions.
Other methods benefiting from deep learning techniques have achieved more accurate and efficient localization.
PoseNet~\cite{kendall2015posenet} and its follow-up works~\cite{walch2017image}, ~\cite{kendall2017geometric}, ~\cite{brahmbhatt2018geometry}, ~\cite{sattler2019understanding} design neural networks to directly predict the pose.
Specially,~\cite{kendall2016modelling}, ~\cite{bui20206d} look into the ambiguity in camera re-localization. 
A Bayesian CNN~\cite{kendall2016modelling} is trained to model the uncertainty in localization, while~\cite{bui20206d,deng2020deep} build a multimodal framework for handling highly ambiguous environments.
Besides one-shot prediction, video-based methods~\cite{zhou2017unsupervised}, ~\cite{clark2017vidloc}, ~\cite{xue2020learning}, ~\cite{radwan2018vlocnet++}, ~\cite{zhou2020kfnet} are introduced to improve temporal consistency of localization. However, these deep learning based methods can easily overfit in the face of small dataset capacity, and struggle to learn useful semantic information in ambiguous scenes.

\begin{figure*}[ht!]
    \centering
    \includegraphics[width=\textwidth]{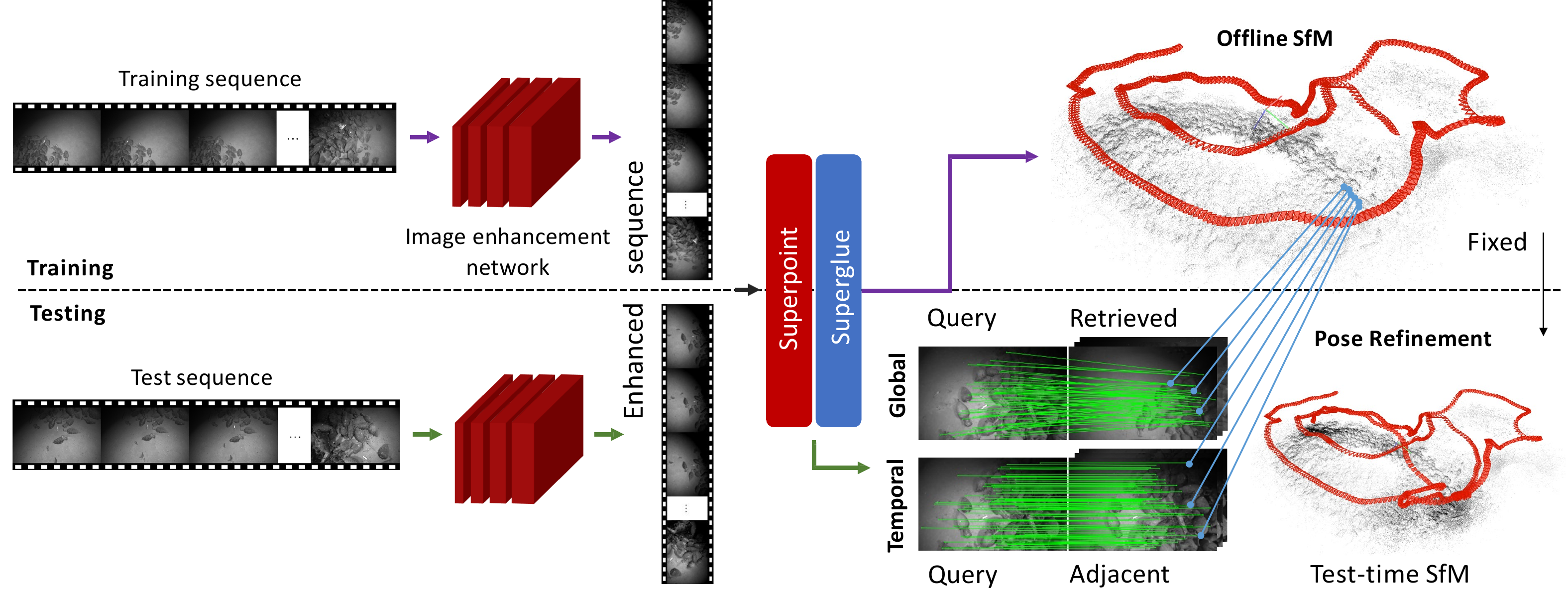}
    \caption{Overview of our system for camera re-localization in extreme environments. Given an image sequence, we train an environment-aware CNN to enhance the degenerated images. A global database can then be established from the training set through an offline SfM reconstruction. 
    For query images, we leverage (a) a global matching with retrieved images and (b) a temporal matching with adjacent frames to find sufficient 2D-3D correspondence.
    A pose refinement is finally performed to optimize the reconstructed poses and the whole 3D structure through incremental SfM.\vspace{-1mm}}
    \label{fig:pipeline}
    \vspace{-15pt}
 \end{figure*}
\vspace{1mm}\noindent\textbf{Correspondence-based relocalization methods.} These methods leverage geometric constraints for reliable pose estimation. 
Scene coordinates~\cite{shotton2013scene} predict 3D coordinates to establish 2D/3D-3D correspondences through a decision tree~\cite{meng2017backtracking,meng2018exploiting,cavallari2019real,dong2021robust}, differential RANSAC~\cite{brachmann2018learning}, ~\cite{brachmann2019expert}, ~\cite{brachmann2021visual} or a voting method~\cite{huang2021vs}.
These methods are accurate but require RGB-D sensors or other auxiliary 3D information such as a pre-scanned model.
Another type of method is built on feature-based reconstruction. 
Visual odometry~\cite{nister2004visual} estimates ego-motion through feature tracking.
Recent advances in visual SLAM~\cite{milford2012seqslam,mur2015orb,mur2017orb} and Struture-from-Motion~\cite{schonberger2016structure} benefit to design more sophisticated systems.
Hierarchical localization~\cite{irschara2009structure,sarlin2019coarse,taira2021video,sarlin21pixloc} combines SfM reconstruction with image retrieval to associate 2D to 3D.
Such methods could fail in extreme conditions, where discriminative features are challenging to detect due to image degeneration, and image retrieval could result in pairs lacking overlap in ambiguous scenarios.

\vspace{1mm}\noindent\textbf{Extreme computer vision.}
Challenging environments often lead to degraded images with low visibility, bad illumination, or ambiguous contents. 
Corrupted vision in adverse weather~\cite{nayar1999vision} has raised wide research interests.
Autonomous driving systems~\cite{lee2018development,bijelic2018benchmarking} handle the challenge by making use of advanced cameras, LiDAR sensors, or GPS for localizing and sensing.
Vision approaches mainly focus on image enhancement. 
Extensive studies have explored fog and haze removal~\cite{he2010single}, ~\cite{cai2016dehazenet}, deraining~\cite{chen2018robust}, deblurring~\cite{kupyn2018deblurgan}, color correction~\cite{li2020underwater},
and domain adaptation~\cite{hoffman2018cycada} between clear scenes and images in adverse conditions. These methods usually rely on hand-crafted priors or synthetic data, which could limit their generalization ability.
Other works pay attention to the lack of challenging datasets. Multimodal datasets~\cite{sattler2018benchmarking}, ~\cite{bijelic2020seeing} provide real world scenes under changing weather, seasons and illumination. 
Recent explorations under the sea~\cite{ferrera2019aqualoc} and extraterrestrial-like environments~\cite{meyer2021madmax} demonstrate extraordinary landscapes with meaningless and ambiguous contents, opening up new space for computer vision to explore. 
Complex visual SLAM algorithms based on depth estimation~\cite{concha2015real} or keyframe detection~\cite{ferrera2019real} are proposed for real-time localization in such environments.\vspace{-2mm}

\section{6D Camera Re-localization in Extreme Environments}
\label{sec:method}\vspace{-2mm}
Our design is motivated by two key challenges: (i) ambiguous elements leading to repetitive patterns, (ii) low image quality often manifested as image blur. To address the former, we propose to leverage temporal information (\cref{sec:temporal enhancement}), which can aid disambiguation. To maximize accuracy, we use the state-of-the-art Superpoint~\cite{detone2018superpoint} and Superglue~\cite{sarlin2020superglue} features as well as a pose refinement module for simultaneously recovering the 3D structure along with the pose. We tackle the latter challenge by prepending our pipeline with an image enhancer (\cref{sec:img enhancement}), which is trained in an end-to-end manner on the downstream task. In what follows, we dive into the details of our algorithm, summarized in~\cref{fig:pipeline}.

\subsection{Preliminaries}
Hierarchical localization~\cite{irschara2009structure,sarlin2019coarse,taira2021video} is a coarse-to-fine framework made for robust and accurate estimation of 6D camera poses. Its two-stage pipeline involves offline database reconstruction and test-time localization. 
The first stage aims to recover the 3D \emph{scene coordinates} from images with known camera poses, while the second stage seeks 2D-3D correspondences w.r.t. this scene reconstruction.

\vspace{1mm}\noindent\textbf{Database construction.}
Given a set of $N$ training images with known poses, a database is constructed to describe the geometry features of the scene:
\begin{equation}
    \mathcal{D}_{train} = \{(I_t, x_t, X_t, g_t) \,|\, 1\le t\le N\}
    \label{eq: database}
\end{equation}
where $I_t\in \mathcal{I}$ is the RGB image, $x_t\in \R^{K\times 2}$ denotes 2D keypoints, whose extraction will be precised later on.
$X_t\in \R^{K\times 3}$ is the 3D coordinates in the world space corresponding to $x_t$, and $g_t\in \R^{d}$ is the $d$-dimensional global descriptor of $I_t$, which encodes the semantic and geometric information of the image.
In practice, such discriminative global descriptors are extracted using a backbone CNN~\cite{arandjelovic2016netvlad}, $f:\mathcal{I}\mapsto\R^d$. 
Our database, $\mathcal{D}_{train}$, can be established by reconstructing the 3D scene through a SfM pipeline~\cite{schonberger2016structure}.

\vspace{1mm}\noindent\textbf{Test-time localization.}
Directly regressing the 6D pose prohibits generalization in localizing a query image $I_q$~\cite{sattler2019understanding}. Hence, we resolve 2D-3D geometry correspondences. Using global descriptors $\{g_t\}$, a set of $N_r$ images potentially sharing covisible parts with $I_q$ are retrieved from the database:
\begin{equation}
    \mathcal{I}_{r_q} = \{(I_{t_i}, x_{t_i}, X_{t_i}, g_{t_i}) \,|\, 1\le i\le N_r\}
    \label{eq: retrieval}
\end{equation}
where $N_r$ is a constant. By matching $I_q$ with the retrieved images on the level of keypoints, we aim to find extensive 2D-3D correspondences. The 6D pose is subsequently estimated through a PnP~\cite{kneip2011novel} algorithm and optimized under a RANSAC scheme~\cite{fischler1981random}.
To this end, in lieu of the classical local descriptors such as SIFT~\cite{lowe2004distinctive}, we opt to obtain the keypoints and their corresponding descriptors via the state-of-the-art Superpoint~\cite{detone2018superpoint} and the corresponding confidences via Superglue~\cite{sarlin2020superglue}. 
This brings efficiency and robustness while improving the matching accuracy, and provides  differentiability to design end-to-end networks.
However, despite the superior performance, the matching can still fail in challenging situations since images with covisible parts can hardly be retrieved due to the ambiguity in the scene.
In addition, the image degeneration in extreme environments (\eg underwater) can further harm keypoint detection and matching.\vspace{-2mm}

\subsection{Localization with Temporal Enhancement}
\label{sec:temporal enhancement}
To handle the ambiguous scenarios, we propose a simple but effective strategy by using temporal information. Incorporating cues from frames along a sequence and their relationships allow for disambiguation while bringing smoothness into the estimation. 

In the spirit of the hierarchical localization~\cite{sarlin2019coarse}, we first match the $N_q$ query images with retrieved images $\{\mathcal{I}_{r_q}\}$.
This global matching step allows us to retrieve a set of frames along with their descriptors. This new database $\mathcal{D}_{query} = \{(I_q, x_q, X_q, g_q)|1\le q\le N_q\}$
contains 2D-3D pairs $x_q$ and $X_q$ which can be contaminated by outliers when images with little covisible content are retrieved.
We then filter out those frames without sufficient 2D-3D correspondences for confident pose estimation, \ie inliers fewer than threshold $s$. 
The remaining $N_k$ images are regarded as \emph{anchor frames} and are considered to be successfully re-localized, \ie $\mathcal{D}_{anchor}\subset \mathcal{D}_{query}$.
We then match query images with their temporal adjacent frames in a sliding window. Suppose that the index $q$ of the query image $I_q$ is sorted according to the timestamp when recorded, given a window size $L$, we match $I_q$ with adjacent frames $\{I_{q_j}\}$ if they are in the anchor database, \ie $q - \frac{L}{2} \le q_j \le q + \frac{L}{2},\, q_j \neq q$ and $I_{q_j}$ is an anchor frame. 
More 2d-3d correspondences can then be constructed since adjacent frames tend to have similar camera poses and large overlapping parts, making keypoint matching much easier.
Newly localized frames with more than $s$ inliers are added to ${D}_{anchor}$, and the temporal matching can be performed again with the updated anchor database. Note that, this paves the way for an iterative process, which increases both the number of matched keypoints and matched frames at each iteration.

\vspace{1mm}\noindent\textbf{Pose refinement.}
Our pipeline ends with a a refinement step to recover all the poses and a complete 3D scene through a test-time SfM reconstruction.
Given the reference SfM model from the training sequence, the anchor frames are registered and triangulation is performed to add 3D points into the scene. This augments the initial SfM model with new points and gives rise to additional 2D-3D correspondences to be used for pose estimation.
In practice, we fix the reference SfM model and the poses of anchor frames. The remaining images are localized through incremental SfM reconstruction.
This step is efficient since the large proportion of frames is expected to be successfully localized after temporal matching, and only a few parameters need to be optimized.

\subsection{Environment-aware Image Enhancement}
\label{sec:img enhancement}
\begin{figure}[t!]
    \includegraphics[width=\linewidth]{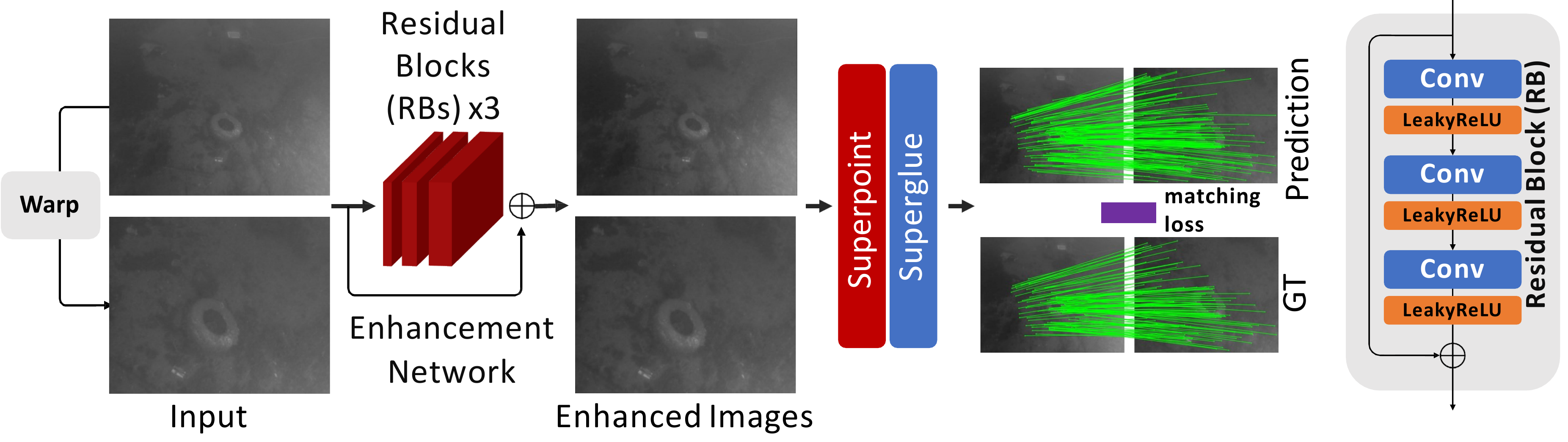}
    \vspace{-15pt}
    \caption{We propose an environment-aware image enhancement method. 
    A CNN is trained to predict the residual of the input image under a self-supervised framework by minimizing the keypoint matching assignment loss.\vspace{-5mm}}
    \label{fig:cnn}
    \vspace{-5pt}
\end{figure}

In addition to the semantic ambiguity, image degeneration can hinder keypoint detection and matching in extreme scenes. This may hamstring the feature-based downstream tasks like SfM reconstruction.
Thus, image enhancement becomes critical to maintain robust localization.
Specially, our goal is to recover the clean latent image $I$ from its possibly degraded version $I^\prime$. A naive approach could use a \emph{denoising autoencoder} predicting $I$ directly. Yet, inspired by~\cite{li2020underwater}, we aim to learn the residual between $I$ \& $I^\prime$:
\begin{equation}
    I = I^\prime + f_e(I^\prime;\theta_e) 
    \label{eq: img_enhancement}
\end{equation}
where $f_e$ is a neural network parameterized by weights $\theta_e$. Typically, $f_e$ is trained on synthesized image pairs $\{(I_i, I_i^{\prime})\}_{i=1}^{N_e}$, where $N_e$ can be infinitely large thanks to the computationally efficient simulation tools~\cite{li2020underwater}. 
However, such a training scheme might lead to a domain gap where the network can end up overfitting to a specific human-designed domain. This either limits real world applications or hampers the final accuracy.
Instead, we design a self-supervised framework as shown in~\cref{fig:cnn}, enabling the enhancement to work for a general purpose image set.

The key to our approach is the use of keypoint correspondences. A clean latent image $I$ tends to have discriminative patterns easily to be detected by local descriptors.
Under the light of this, we pose our task to be the discovery of a latent image $I$ which maximizes the detection of discriminative features and the accuracy of keypoint matching. 
To this end, we train the network to recover $I$ by minimizing the keypoint assignment loss. 
Specially, for a possibly degenerated input image $I_1^\prime$, we first randomly draw homography parameters $\mathcal{H}$ and warp $I_1^\prime$ to obtain $I_2^\prime$.
The enhanced image pair $(I_1,I_2)$ is then fed to Superpoint~\cite{detone2018superpoint} to detect the keypoint sets $\mathcal{A}$ and $\mathcal{B}$ on images $I_1$ and $I_2$, respectively:
\begin{equation}
    \mathcal{A} = \{(x_i, h_i)\,|\,1\le i \le N_1\}, 
    \mathcal{B} = \{(x_j, h_j)\,|\,1\le j \le N_2\}
    \label{eq: keypoints}
\end{equation}
Here, $x$ is the 2D keypoint, and $h$ refers to its local descriptor. 
Based on $x$ and $h$, Superglue~\cite{sarlin2020superglue} predicts the matching probability $p$ between each keypoint pair.
By defining keypoint index sets $\mathcal{K}_1=\{i\,|\,1\le i \le N_1\}$ and $\mathcal{K}_2=\{i\,|\,1\le i \le N_2\}$, we generate the ground truth matches $\mathcal{M} = \{(i, j)\} \subset \mathcal{K}_1 \times \mathcal{K}_2$ and unmatched keypoints 
$\mathcal{I}\subset \mathcal{K}_1$, $\mathcal{J}\subset \mathcal{K}_2$ using the homography $\mathcal{H}$. 
Our loss function follows Superglue~\cite{sarlin2020superglue}:
\begin{equation}
    \mathcal{L}_m = -\sum_{(i, j) \in \mathcal{M}}\log p_{i,j} - \sum_{i \in \mathcal{I}}\log p_{i,N_2+1} 
                - \sum_{j \in \mathcal{J}}\log p_{N_1+1, j}
    \label{eq: loss_match}
\end{equation}
where $p_{i, j}$ is the assignment probability predicted by Superglue, and $p_{i,N_2+1}$ and $p_{N_1+1, j}$ refer to the probability of unmatched keypoints. The loss aims to maximize matching precision and recall, guiding the network to recover latent clear images with discriminative features. Practically we fix Superglue and Superpoint to stabilize training.

In addition to the keypoint assignment loss, we further use a smoothness term to prevent trivial solutions:
\begin{equation}
    \mathcal{L}_s = \frac{1}{N}||I_1-I^\prime_1||_2^2
    \label{eq: loss smooth}
\end{equation}
where $N$ is the number of pixels. Combining~\cref{eq: loss_match} and ~\cref{eq: loss smooth}, we train our network end-to-end to minimize $\mathcal{L} = \mathcal{L}_m + \lambda \mathcal{L}_s$, where $\lambda$ is for balancing the two losses.
\section{Experiments and evaluations}
\label{sec:exp}

In this part, we explain our system implementation and experimental design. Our high level goal is to answer the following core questions: 
\begin{enumerate}[(a)]
    \item Does our method perform well for highly ambiguous extreme environments?
    \item Does our design hurt its performance in non-extreme environments? 
    \item What is the contribution of each part of the algorithm to the final performance?
\end{enumerate}
We then find the answers through extensive experiments.

\begin{figure*}[ht!]
    \centering
    \includegraphics[width=\textwidth]{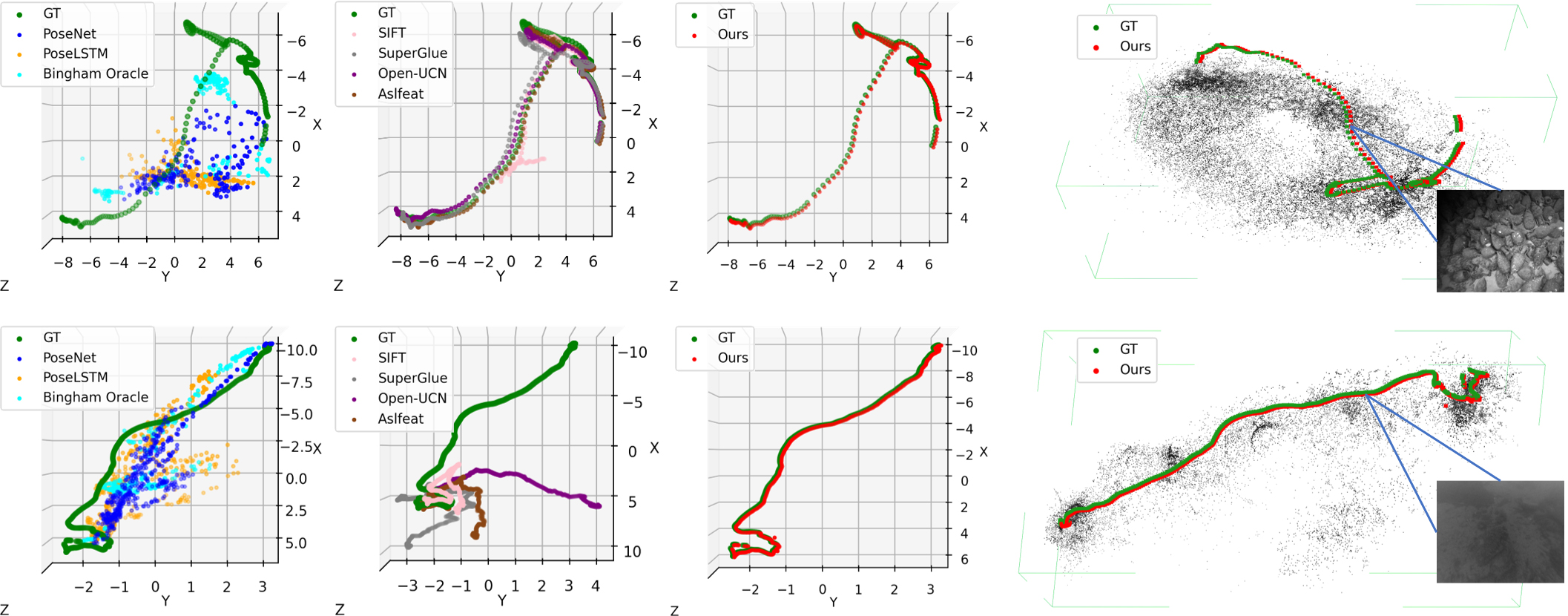}
    \vspace{-10pt}
    \caption{Camera trajectory of the baseline methods and ours on archaeological sequence 10 where images are filled with repetitive patterns (top) and harbor sequence 2 with blurred images (bottom) from Aqualoc dataset.}
    \label{fig:trajectory}
 \end{figure*}
\vspace{1mm}\noindent\textbf{Implementation details.} 
To build the database from the training sequence, we utilize pretrained Superpoint~\cite{detone2018superpoint} and Superglue~\cite{sarlin2020superglue} to extract and track sparse features, where up to 512 discriminative keypoints in each frame are detected and matched for efficient SfM reconstruction~\cite{schonberger2016structure}.
Instead of matching image pairs exhaustively where the complexity can be $\mathcal{O}(n^2)$ with $n$ frames, we match each image with its temporally adjacent 50 frames to accelerate the database establishment.
For the global descriptor, we refer to the state-of-the-art image retrieval method SFRS~\cite{ge2020self}.
During the inference time, $N_r=30$ images potentially overlapped with the query image are retrieved from the database for the global matching step. Temporal matching is then performed within a 30 frame window (\ie window size $L=30$) for 10 iterations, and we empirically define an inlier threshold $s=50$. 
After that, the whole 3D scene and camera poses are optimized in the pose refinement step through incremental SfM reconstruction, where we fix the offline-constructed SfM model and poses of anchor frames (inliers more than $s$). For image enhancement, we design a lightweight network consisting of 3 residual blocks with 64 feature channels in the intermediate layers (\cref{fig:cnn}). We train the network for 10 epochs with $1\times 10^{-4}$ learning rate to minimize the loss $\mathcal{L}$ with $\lambda=1$.
Note that we skip this step on 7-Scenes dataset since there is no need to enhance already high-quality indoor images. 

\begin{table}[b!]
    \caption{Average results of 10 scenes from Aqualoc dataset. $*$ means failure cases exist and we only report successful results (here COLMAP succeeds on 6 out of 10 sequences). \emph{Ours (vanilla)} indicates without refined or enhanced. 
    \emph{Ours (finetuned)} denotes we finetune Superpoint and Superglue instead of image enhancement.}
    \label{tab:aqualoc}
    \begin{center}
        \resizebox{0.7\linewidth}{!}{
        \begin{tabular}{lcc|cc}
            \toprule
            \textbf{Model} & \textbf{0.1m, 5$^{\circ}$}$\uparrow$ & \textbf{0.5m, 15$^{\circ}$}$\uparrow$ 
             & \makecell[c]{\textbf{Translation(m)}$\downarrow$\\Avg  $\pm$  std} & \makecell[c]{\textbf{Rotation($^{\circ}$)}$\downarrow$\\Avg $\pm$ std} \\  
            \midrule 	
            PoseNet~\cite{kendall2015posenet} & 0.27\% & 4.81\% & 3.10 $\pm$ 1.92 & 86.13 $\pm$ 46.031\\
            PoseLSTM~\cite{walch2017image} & 0.26\% & 4.91\% & 2.92 $\pm$ 1.55 & 83.84 $\pm$ 46.18\\
            Bingham~\cite{bui20206d} & 1.62\% & 4.74\% & 4.02 $\pm$ 2.80 & 88.57 $\pm$ 34.49\\
            Bingham (Oracle)~\cite{bui20206d} & 2.72\% & 5.90\% & 2.22 $\pm$ 1.18 & 35.85 $\pm$ 12.05\\
            SfM Learner~\cite{zhou2017unsupervised} & 0.56\% & 2.91\% & 4.54 $\pm$ 0.38 & 76.98 $\pm$ 24.29\\
            \midrule
            DSAC++~\cite{brachmann2018learning} &0.24\% & 0.24\% & 10.78 $\pm$ 0.61 & 81.89 $\pm$ 43.68\\
            ESAC~\cite{brachmann2019expert} & 0.66\% & 1.81\% & 8.95 $\pm$ 0.72 & 90.64 $\pm$ 24.90\\
            DSAC*~\cite{brachmann2021visual} & 0.10\% & 1.19\% & 9.53 $\pm$ 0.53 & 69.10 $\pm$ 22.10\\
            HFNet~\cite{sarlin2019coarse} & 32.33\% & 33.39\% & 3.30 $\pm$ 2.39 & 53.34 $\pm$ 23.10\\
            PixLoc~\cite{sarlin21pixloc} & 14.78\% & 16.39\% & 1.10 $\pm$ 0.31 & 84.05 $\pm$ 49.15\\
            \midrule 
            SIFT~\cite{lowe2004distinctive} & 16.65\% & 34.85\% & 2.35 $\pm$ 3.20 & 51.16 $\pm$ 47.77 \\
            SURF~\cite{bay2006surf} & 11.27\% & 36.84\% & 1.78 $\pm$ 2.61 & 41.49 $\pm$ 39.19\\
            Superglue~\cite{sarlin2020superglue} &9.05\% & 26.64\% & 2.03 $\pm$ 2.92 & 40.19 $\pm$ 31.06\\
            Open-UCN~\cite{choy2016universal} &11.12\% & 31.83\% & 1.68 $\pm$ 2.01 & 28.81 $\pm$ 22.22\\
            ASLFeat~\cite{luo2020aslfeat} & 10.47\% & 32.05\% & 1.64 $\pm$ 1.92 & 33.08 $\pm$ 31.54\\
            COLMAP~\cite{schonberger2016structure} & \emph{72.95\%}$^*$ & \emph{100\%}$^*$ & \emph{0.09} $\pm$ \emph{0.06}$^*$ & \emph{1.01} $\pm$ \emph{0.58}$^*$\\
            \midrule
            Ours \footnotesize{(vanilla)} & 53.69\% & 59.37\% & 0.92 $\pm$ 1.14 & 6.97 $\pm$ 9.09\\
            Ours \footnotesize{(w/o refined)} & 73.75\% & 74.76\% & 0.50 $\pm$ 0.77 & 4.29 $\pm$ 8.53\\
            Ours \footnotesize{(w/o enhanced)} & 71.97\% & 85.90\% & 0.17 $\pm$ 0.32 & 1.63 $\pm$ 2.88\\
            Ours \footnotesize{(finetuned)} & 74.30\% & 94.62\% & 0.18 $\pm$ 0.22 & 1.68 $\pm$ 2.77 \\
            Ours (full) & \textbf{94.79}\% & \textbf{98.51\%} & \textbf{0.02} $\pm$ 0.01 & \textbf{0.37} $\pm$ 0.21\\
            \bottomrule
    \end{tabular}}
    \end{center}
    \vspace{-20pt}
\end{table}

\vspace{1mm}\noindent\textbf{Baselines \& Datasets.}
\label{sec:baselines}
We train and test the direct pose estimation methods and correspondence based approaches.
For the first category, we refer to PoseNet~\cite{kendall2015posenet}, PoseLSTM~\cite{walch2017image}, Bingham~\cite{bui20206d}, and SfM Learner~\cite{zhou2017unsupervised}. 
The second type includes DSAC++~\cite{brachmann2018learning}, ESAC~\cite{brachmann2019expert}, DSAC*~\cite{brachmann2021visual} based on differential RANSAC, and HF-Net~\cite{sarlin2019coarse}, Pixloc~\cite{sarlin21pixloc} using hierarchical localization and image retrieval. 
For the above methods, the pretrained weights are finetuned on the target sequences except that the pretrained models are used for SfM reconstruction in HF-Net and Pixloc. 
To further explore the factors that could affect the performance of feature-based localization systems, we estimate visual odometry using different descriptors and matching algorithms, \ie classical methods including SIFT~\cite{lowe2004distinctive} and SURF~\cite{bay2006surf}, 
learning based methods including Open-UCN~\cite{choy2016universal}, ASLFeat~\cite{luo2020aslfeat} and Superglue~\cite{sarlin2020superglue}. 
For simplification, we decompose relative poses between continuous frames from 2D-2D correspondences and use ground truth to scale the translation.
Additionally, as a simple baseline of our system, we use COLMAP~\cite{schonberger2016structure} to construct a local map on the test sequence, and merge it with the offline-constructed SfM model based on correspondences built from the global matching step as described in \cref{sec:temporal enhancement}. 

To evaluate our method, we choose two visually ambiguous datasets and one standard dataset to benchmark the proposed method.

\begin{figure}[t!]
    \centering
    \subfigure[Camera trajectory]{\includegraphics[width=0.35\linewidth]{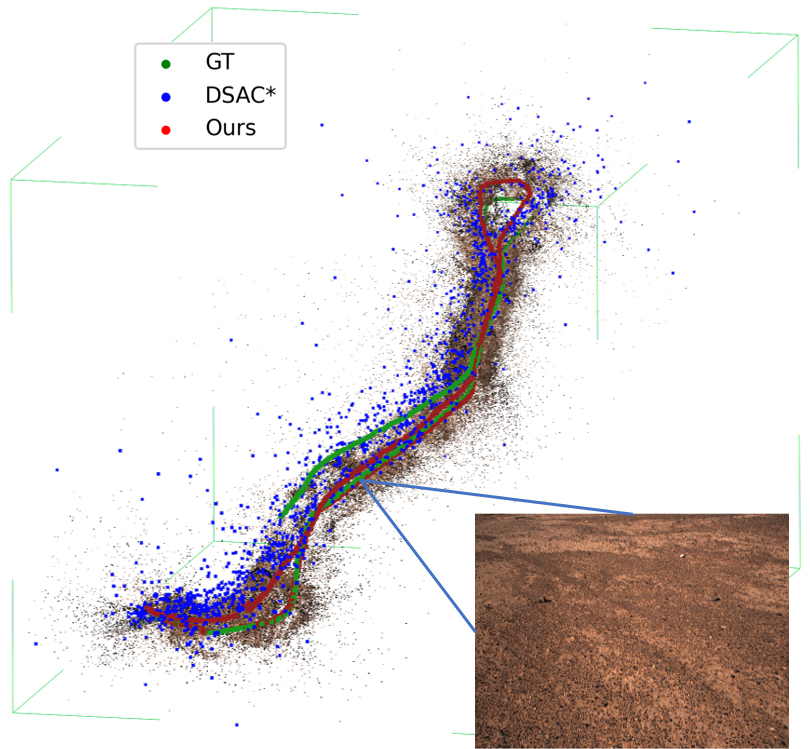}}
    \subfigure[Aligned with IMU data]{\includegraphics[width=0.35\linewidth]{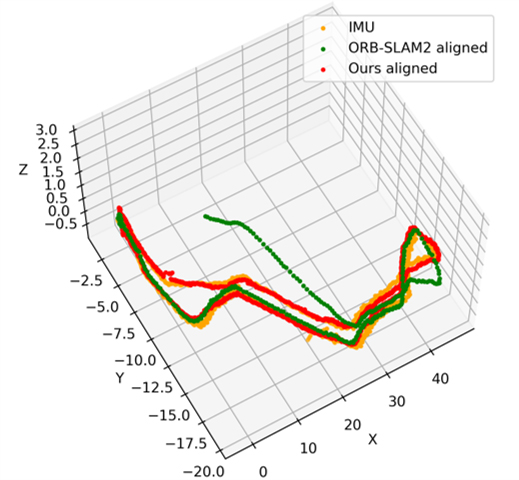}}
    \vspace{-5pt}
    \caption{\textbf{(a)} 3D scene reconstruction and camera trajectory on Mars-Analogue F location. 
    \textbf{(b)} Compared to the provided poses from ORB-SLAM2~\cite{mur2017orb} which suffers from drift effects, our results are better aligned with the trajectory from IMU signals, indicating the robustness of our system.}
    \label{fig:trajectory_mars}
    \vspace{-10pt}
\end{figure}
\vspace{1mm}\noindent\textbf{Aqualoc dataset}~\cite{ferrera2019aqualoc} is an underwater dataset made for relocalization purposes.
The dataset contains videos collected from a harbor site at a shallow sea and an archaeological site in the deep sea.  
The images are poorly illuminated, blurred, or filled with ambiguous contents like repetitive patterns.
Camera poses calculated from COLMAP~\cite{schonberger2016structure} are provided as ground truth.
We choose 5 self-contained scenes from the archaeological site and another 5 from the harbor site, with averagely 542 images for training and 183 images for evaluation in each scene.

\vspace{1mm}\noindent\textbf{Mars-Analogue dataset}~\cite{meyer2021madmax} is acquired from Mars analog sites in Morocco desert.
The scenes are distinguished by desolate landscapes such as sands and rocks, where these textureless places and repetitive contents exacerbate ambiguity in localization. We use the provided camera poses estimated from ORB-SLAM2~\cite{mur2017orb} as ground truth and choose 3 scenes collected from 3 different locations with on average 1659 images for training and 1542 images for testing.

\vspace{1mm}\noindent\textbf{7-Scenes dataset}~\cite{shotton2013scene} consists of 7 indoor scenarios collected by Kinect RGB-D cameras, and we use only RGB data in our experiments.
Pictures taken from such indoor environments offer sufficient visual cues to infer the poses. We use this dataset to sanity check that our method does not cause deteriorated results for common daily scenes.

\begin{table}[ht!]
    \caption{Results of 3 scenes from Mars-Analogue dataset.}
    \label{tab:mars}
    \begin{center}
        \resizebox{0.7\linewidth}{!}{
        \begin{tabular}{lcc|cc}
            \toprule
            \textbf{Model} & \textbf{0.1m, 5$^{\circ}$}$\uparrow$ & \textbf{0.5m, 15$^{\circ}$}$\uparrow$ 
             & \makecell[c]{\textbf{Translation(m)}$\downarrow$\\Avg  $\pm$  std} & \makecell[c]{\textbf{Rotation($^{\circ}$)}$\downarrow$\\Avg $\pm$ std} \\
            \midrule 	
            PoseNet~\cite{kendall2015posenet} & 0.32\% & 10.01\% & 3.38 $\pm$ 1.73 & 13.21 $\pm$ 4.63\\
            PoseLSTM~\cite{walch2017image} & 0.15\% & 8.45\% & 4.14 $\pm$ 2.41 & 12.77 $\pm$ 3.23\\
            Bingham~\cite{bui20206d} & 2.08\% & 8.98\% & 2.20 $\pm$ 0.92& 12.90 $\pm$ 4.86\\
            Bingham (Oracle)~\cite{bui20206d} & 2.31\% & 15.51\% & 2.06 $\pm$ 0.74& 8.17 $\pm$ 2.44\\
            SfM Learner~\cite{zhou2017unsupervised} & 2.24\% & 8.54\% & 3.88 $\pm$ 1.52 & 60.83 $\pm$ 21.06\\
            \midrule
            DSAC++~\cite{brachmann2018learning} & 0.15\% & 4.54\% & 7.35 $\pm$ 3.36& 12.82 $\pm$ 2.99\\ 
            ESAC~\cite{brachmann2019expert} & 0.74\% & 9.29\% & 4.10 $\pm$ 1.58& 12.39 $\pm$ 1.60\\
            DSAC*~\cite{brachmann2021visual} & 7.15\% & 20.94\% & 2.34 $\pm$ 0.94& 11.10 $\pm$ 3.75\\ 
            HFNet~\cite{sarlin2019coarse} & 27.97\% & 55.50\% & 0.99 $\pm$ 0.45 & 5.70 $\pm$ 2.31\\
            PixLoc~\cite{sarlin21pixloc} & 7.40\% & 8.67\% & 3.13 $\pm$ 1.74 & 14.90 $\pm$ 4.94\\
            \midrule 
            SIFT~\cite{lowe2004distinctive} & 19.24\% & 33.44\% & 1.72 $\pm$ 1.06& 22.04 $\pm$ 12.76\\
            SURF~\cite{bay2006surf} & 14.30\% & 26.62\% & 2.63 $\pm$ 1.38& 35.45 $\pm$ 19.15\\
            Superglue~\cite{sarlin2020superglue} &7.19\% & 25.67\% & 2.16 $\pm$ 1.21& 50.55 $\pm$ 33.22\\
            Open-UCN~\cite{choy2016universal} &7.04\% & 11.21\% & 6.43 $\pm$ 2.56& 103.90 $\pm$ 12.57\\
            ASLFeat~\cite{luo2020aslfeat} & 7.77\% & 17.64\% & 3.72 $\pm$ 1.35& 64.32 $\pm$ 16.91\\
            \midrule
            Ours \footnotesize{(vanilla)} & 29.47\% & 57.62\% & 0.90 $\pm$ 0.46& 5.61 $\pm$ 2.27\\
            Ours \footnotesize{(w/o refined)} & 30.03\% & 55.16\% & 0.80 $\pm$ 0.50& 5.66 $\pm$ 2.35\\
            Ours \footnotesize{(w/o enhanced)} & 31.41\% & 59.53\% & 0.83 $\pm$ 0.48& 5.57 $\pm$ 2.21\\
            Ours \footnotesize({full)} & \textbf{32.42}\% & \textbf{59.60\%} & \textbf{0.79} $\pm$ 0.50& \textbf{5.56} $\pm$ 2.25\\
            \bottomrule
    \end{tabular}}
    \end{center}
    \vspace{-20pt}
\end{table}

\subsection{Evaluation \& Results}
In the following, we try to answer question (a), (b), and (c) introduced at the beginning of the section.

\vspace{1mm}\noindent\textbf{Evaluation Metrics.} 
We report the median rotation and translation errors along with standard deviations on each dataset. We also show the percentage of correctly localized frames under error thresholds chosen roughly proportional to the scale of the scene. 

\begin{table}[ht!]
    \caption{Results on 7-Scenes, where dash ($-$) means that the particular result was not reported. Our method achieves comparative performance and maintains robust with less training data (\emph{Ours x\%} means using x\% images for training).}
    \vspace{-5pt}
    \label{tab:7scenes}
    \begin{center}
        \resizebox{0.55\linewidth}{!}{
        \begin{tabular}{lc|cc}
            \toprule
            \textbf{Model} & \textbf{5cm, 5$^{\circ}$}$\uparrow$ 
             & \makecell[c]{\textbf{Translation(cm)}$\downarrow$\\Avg$\pm$std} & \makecell[c]{\textbf{Rotation($^{\circ}$)}$\downarrow$\\Avg$\pm$std} \\  
            \midrule 	
            PoseNet~\cite{kendall2015posenet} & $-$ & 44.10 $\pm$ 9.54 & 5.24 $\pm$ 1.37\\
            PoseLSTM~\cite{walch2017image} & $-$ & 31.29 $\pm$ 6.32 & 9.85 $\pm$ 2.99\\
            Bingham~\cite{bui20206d} & $-$ & 20.43 $\pm$ 7.11 & 8.64 $\pm$ 2.81\\
            Vidloc~\cite{clark2017vidloc} & $-$ & 24.57 $\pm$ 7.01 & $-$ \\
            VdlocNet++~\cite{radwan2018vlocnet++} & $-$ & \textbf{2.16} $\pm$ 0.32 & 1.39 $\pm$ 0.40 \\
            \midrule
            DSAC++~\cite{brachmann2018learning} & 60.40\% & 8.43 $\pm$ 8.94 & 2.40 $\pm$ 2.26\\
            ESAC~\cite{brachmann2019expert} & 73.80\% & 3.40 $\pm\,\,\,\,-\,\,\,$ & 1.50 $\pm\,\,\,\,-\,\,\,$\\
            DSAC*~\cite{brachmann2021visual} & \textbf{80.70}\% & 2.69 $\pm$ 1.08 & 1.41 $\pm$ 0.25\\
            HF-Net~\cite{sarlin2019coarse} & $-$ & 4.19 $\pm$ 1.91 & 1.37 $\pm$ 0.37\\
            PixLoc~\cite{sarlin21pixloc} & 75.7\% & 2.86 $\pm$ 1.25 & \textbf{0.98} $\pm$ 0.22\\
            \midrule
            Ours (5\%) & 69.09\%& 3.92 $\pm$ 1.92 & 1.24 $\pm$ 0.35\\
            Ours (20\%) & 76.14\%& 3.30 $\pm$ 1.30 & 1.09 $\pm$ 0.22\\
            Ours (100\%) & 76.53\%& 3.19 $\pm$ 1.37 & 1.02 $\pm$ 0.25\\
            \bottomrule
    \end{tabular}}
    \end{center}
    \vspace{-20pt}
\end{table}
\vspace{1mm}\noindent\textbf{Performance on Aqualoc dataset.} 
The quantitative results can be found in~\cref{tab:aqualoc}. 
Current approaches are not effective enough to handle ambiguous underwater scenes.
As shown in~\cref{tab:aqualoc}, the state-of-the-art method Bingham~\cite{bui20206d} designed to model the ambiguity clearly fails even if it chooses the closest pose to the ground truth among multi-hypotheses (\emph{Bingham Oracle}). Simply incorporating temporal information such as visual odometry estimation or using COLMAP~\cite{schonberger2016structure} (described in \cref{sec:baselines}) significantly boosts accuracy, but could fail in some challenging cases, \eg COLMAP failing on 4 sequences out of 10. 
Qualitative results in~\cref{fig:trajectory} illustrate that visual odometry estimation performs well on the archaeological sequence with clear images but fails on the harbor sequence where images are degenerated.
Our method can re-localize nearly all the frames within small errors.

\vspace{1mm}\noindent\textbf{Performance on Mars-Analogue dataset.} 
Quantitative results are reported in~\cref{tab:mars}. 
Compared to the underwater Aqualoc dataset, images from the Mars-Analogue dataset are clearer and high-quality, which sheds light on the better performance of baseline methods. 
However, the visual odometry estimation struggles to handle the common textureless regions in the desert environment. COLMAP baseline totally fails on this dataset, and thus its results are not reported.
~\cref{fig:trajectory_mars} (a) shows that our system manages to generate a coherent trajectory, while results of the state-of-the-art method DSAC*~\cite{brachmann2021visual} are noisy due to the ambiguity. 
In addition, we argue that the provided camera poses from ORB-SLAM2~\cite{mur2017orb} may suffer from drift effects as shown in~\cref{fig:trajectory_mars} (b), and thus the quantitative performance of our method is not so impressive.
However, even though we use the potentially inaccurate poses for training, qualitative results in~\cref{fig:trajectory_mars} (b) imply that our results are better aligned with trajectory calculated from IMU signals, further demonstrating the robustness of our system.

\begin{figure}[t!]
    \centering
    \subfigure[Raw image]{\includegraphics[width=0.35\linewidth]{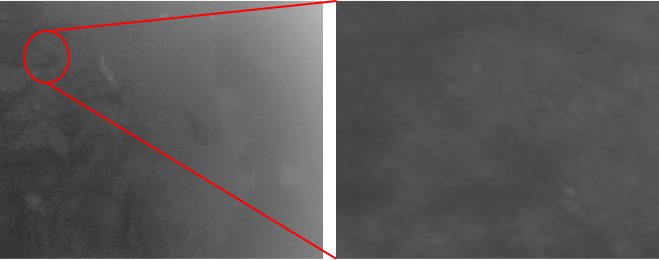}}
    \subfigure[Enhanced image]{\includegraphics[width=0.35\linewidth]{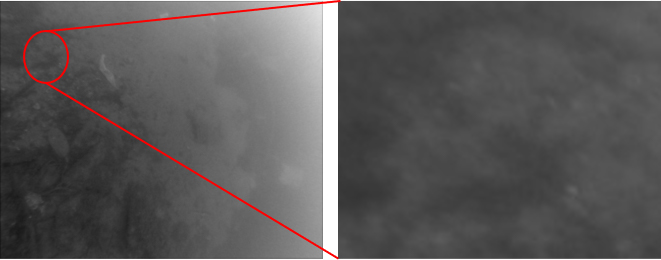}}\vspace{-5pt}

    \subfigure[Raw pair matching]{\includegraphics[width=0.35\linewidth]{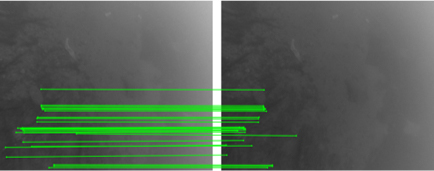}}
    \subfigure[Enhanced pair matching]{\includegraphics[width=0.35\linewidth]{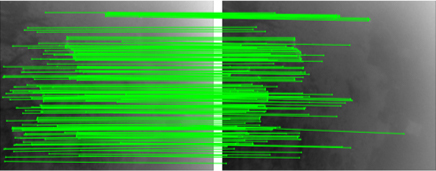}}\vspace{-5pt}
    
    \subfigure[Results without enhancement]{\includegraphics[width=0.35\linewidth]{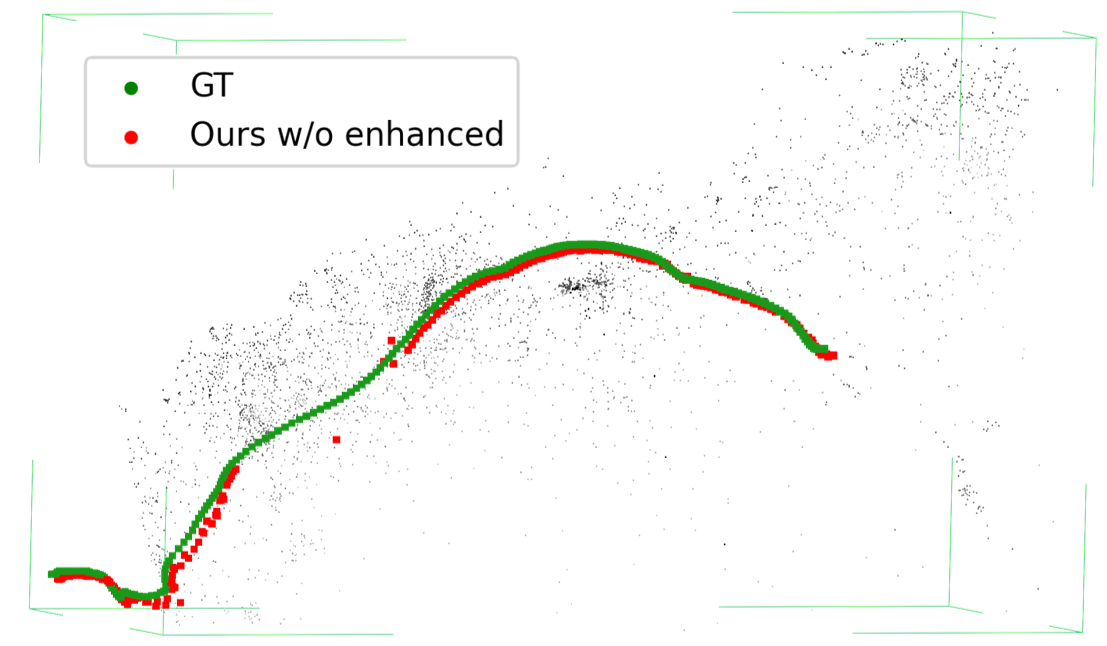}}
    \subfigure[Results with enhancement]{\includegraphics[width=0.35\linewidth]{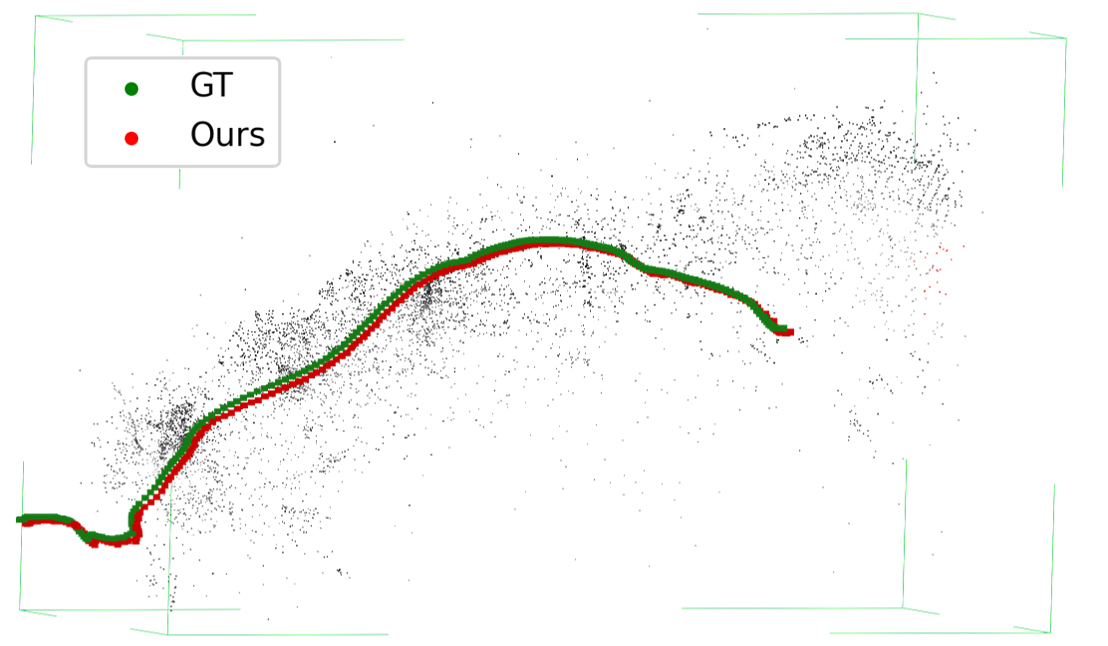}}
    \vspace{-5pt}
    \caption{The environment-aware image enhancement method manages to improve the image quality (top) which leads to better matching (middle), denser points in SfM reconstruction and more coherent localization results (bottom).}
    \label{fig:enhancement}
    \vspace{-5pt}
\end{figure}
\begin{figure}[ht!]
    \centering
    \subfigure[Ours 5\%: 84.95\% ($5cm, 5^\circ$)]{\includegraphics[width=0.4\linewidth]{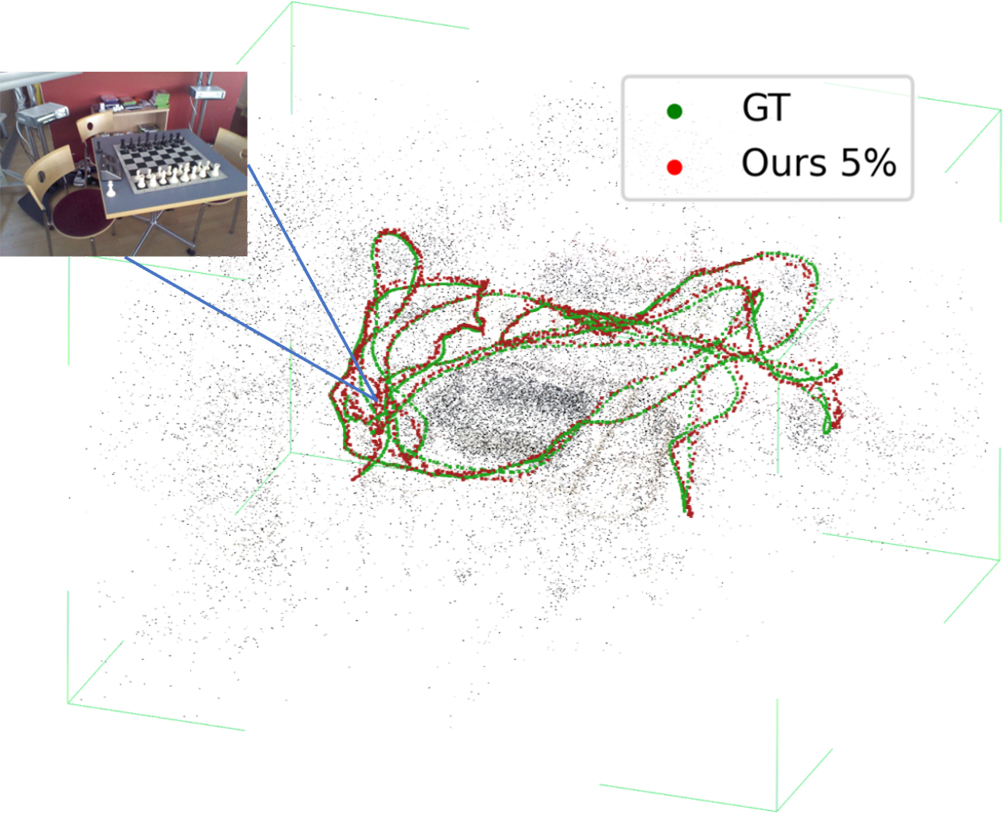}}
    \subfigure[Ours 100\%: 93.45\% ($5cm, 5^\circ$)]{\includegraphics[width=0.4\linewidth]{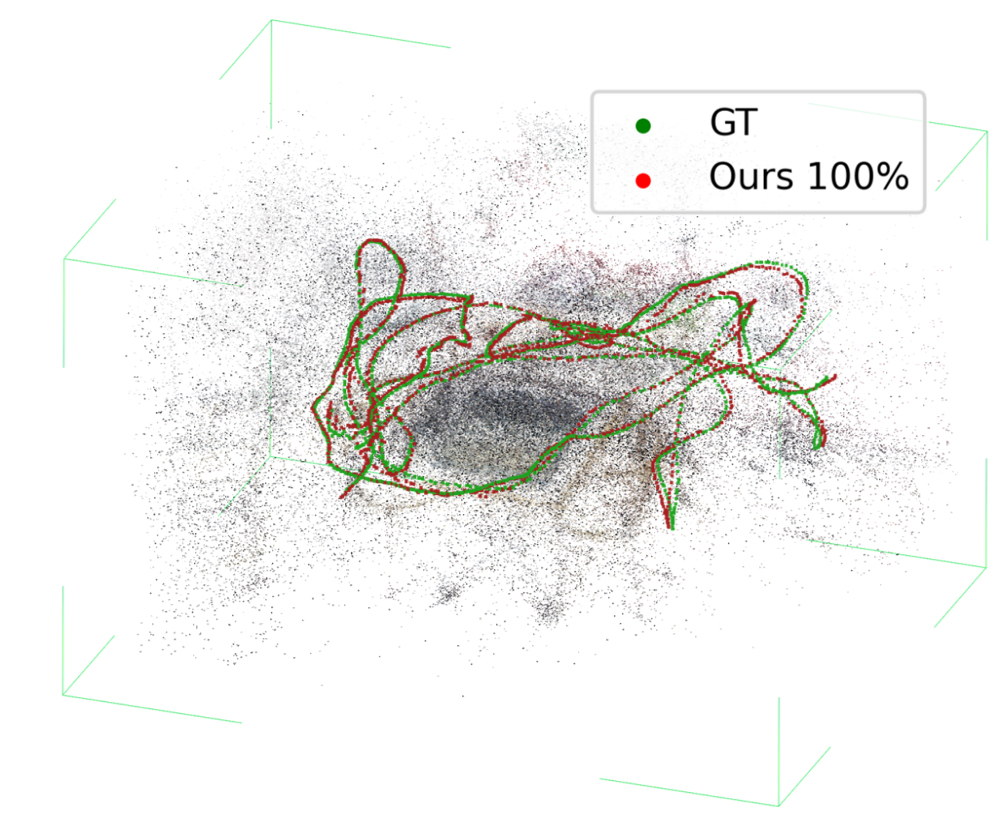}}
    \vspace{-5pt}
    \caption{On Chess scene from 7-Scenes dataset, our system remains robust and accurate using only 5\% training data.\vspace{-3mm}}
    \label{fig:training data}
    \vspace{-5pt}
\end{figure}

\vspace{1mm}\noindent\textbf{Performance on 7-Scenes dataset.} Quantitative results in~\cref{tab:7scenes} show that our method achieves comparative performance on 7-Scenes compared with the state-of-the-art approaches.
Since the hierarchical localization pipeline is designed for large-scale environments, our method might not have advantages in fine-grained reconstruction and localization.
However, the dense indoor image data might be redundant for our system to recover the sparse 3D scene. 
When using fewer images for training, our method maintains robust localization as demonstrated in~\cref{tab:7scenes}.
\cref{fig:training data} shows that using only 5\% training data, re-localization results of our system can still be reasonable and accurate.
\newcommand{\specialcell}[2][c]{%
  \begin{tabular}[#1]{@{}c@{}}#2\end{tabular}}
  
\begin{table}[ht!]
    \vspace{-10pt}
    \setlength\tabcolsep{1em}
    \caption{Accuracy of our method implemented with different discriptors on the archaeological sequence 10 (without enhancement) from Aqualoc dataset.}
    \label{tab:abaltion_kpt}
    \vspace{-5pt}
    \begin{center}
        \resizebox{\linewidth}{!}{
        \begin{tabular}{lccc}
            \toprule
             & SIFT~\cite{lowe2004distinctive} & ASLFeat~\cite{luo2020aslfeat} & Superpoint~\cite{detone2018superpoint} \& Superglue~\cite{sarlin2020superglue}\\
            \midrule
            
            Translation(m)$\downarrow$ & 1.60 & 0.05 & 0.01\\
            Rotation($^{\circ}$)$\downarrow$ & 20.56 & 0.59 & 0.32\\
            0.1m, 5$^{\circ}$$\uparrow$ & 27.88\% & 78.85\% & 86.54\%\\
            \bottomrule
    \end{tabular}}
    \end{center}
    \vspace{-20pt}
\end{table}

\subsection{Ablation Studies}
The section aims to answer question (c) by finding the factors that contribute to the performance of our system.

\vspace{1mm}\noindent\textbf{Use of temporal information in matching and pose refinement.}
Without temporal information, our method is similar to HF-Net~\cite{sarlin2019coarse} which could generate noisy results when failing to retrieve images with covisible parts in ambiguous situations.
To address the issue, we propose a temporal matching and pose refinement method introduced in~\cref{sec:temporal enhancement}. 
Results in~\cref{tab:aqualoc} and~\cref{tab:mars} indicate the effectiveness of the design, where \emph{Ours (vanilla)} denotes our method with only temporal matching and \emph{Ours (w/o refined)} means ours without pose refinement.
Both variants display much better accuracy than HF-Net, implying how temporal matching can disambiguate localization.
But there is clearly a gap between the full design, where the localization results and the whole 3D structure are optimized through pose refinement.

\begin{figure}[t!]
    \centering
    \subfigure[Ours 100\%: 53.20\% ($5cm, 5^\circ$)]{\includegraphics[width=0.4\linewidth]{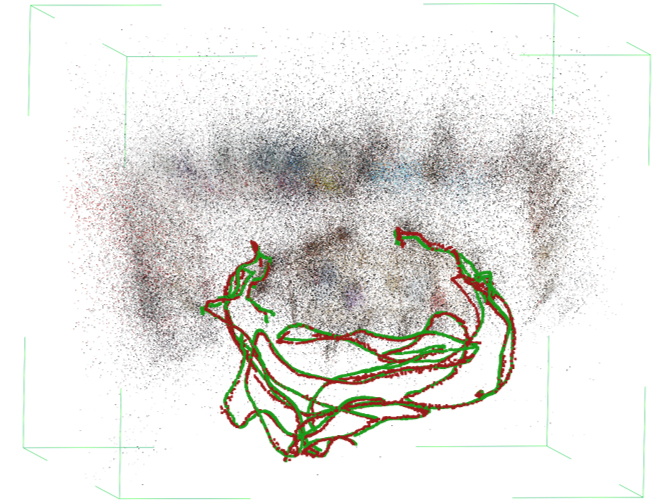}}
    \subfigure[Reprojection of reconstruction]{\includegraphics[width=0.4\linewidth]{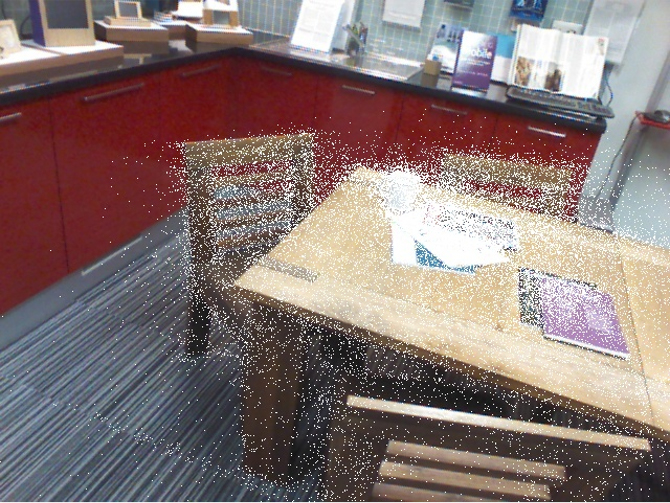}}
    \vspace{-5pt}
    \caption{Performance of our method on Kitchen scene from 7-Scenes. In (b) we show the reprojection of reconstruction 3D points (chairs and tables). 
    The noise in SfM reconstruction results and camera intrinsic parameters can lead to loss in localization accuracy, which is a limitation of our method.}
    \label{fig:limitation}
    \vspace{-5pt}
\end{figure}
\vspace{1mm}\noindent\textbf{Use of image enhancement.}
We propose an environment-aware image enhancement method (~\cref{sec:img enhancement}) to improve downstream feature-based reconstruction.
In~\cref{fig:enhancement}, the method successfully recovers the latent images from blurred inputs, enabling to improve keypoint matching which leads to denser reconstruction and better localization results.
~\cref{tab:aqualoc} and~\cref{tab:mars} show loss in accuracy of our method without image enhancement (\emph{Ours(w/o enhanced)}).\begin{table}[ht!]
    \caption{Average time per frame (s) for each step of HF-Net~\cite{sarlin2019coarse} and our method on Aqualoc dataset~\cite{ferrera2019aqualoc} ($-$: not reported, $/$: not required).}
    \label{tab:time}
    \vspace{-5pt}
    \begin{center}
        \resizebox{0.65\linewidth}{!}{
        \begin{tabular}{cccccc}
            \toprule
             & \specialcell{Offline \\database building} & \specialcell{Image \\ retrieval} & \specialcell{Global\\ matching} & \specialcell{Temporal \\ matching} & \specialcell{Test-time \\ SfM} \\
            \midrule
            HF-Net~\cite{sarlin2019coarse} & $-$ & 0.02 & 0.04  & $/$ & $/$ \\
            Ours & 7.0 & 0.04 & 2.40 & 2.65 & 0.34 \\
            \bottomrule
    \end{tabular}}
    \end{center}
    \vspace{-20pt}
\end{table}

Additionally, we find finetuning the pretrained Superpoint and Superglue on the target domain can only provide limited improvement (\emph{ours (finetuned)} in \cref{tab:aqualoc}), showing the necessity to enhance the degenerated images.
However, on Mars-Analogue dataset, the improvements brought by image enhancement can be marginal since images are already in high quality.

\vspace{1mm}\noindent\textbf{Use of Superpoint and Superglue.} We take advantage of the state-of-the-art methods Superpoint~\cite{detone2018superpoint} and Superglue~\cite{sarlin2020superglue} for feature detection and tracking to attain the best accuracy. \cref{tab:abaltion_kpt} shows that other methods including classical descriptors such as SIFT~\cite{lowe2004distinctive} and learned features like ASLFeat~\cite{luo2020aslfeat} could hamper the performance.
\vspace{-1mm}\subsection{Discussions and Limitations}\vspace{-1mm}
\cref{tab:time} presents average timings of our pipeline, which is not achieving real-time as is with a single-core implementation. 
The majority of the time is consumed by the 2D keypoints matching step in both global matching and temporal matching, which can be significantly reduced in a parallel manner. Simple implementing the system with 12 cores will improve the processing speed to 1.25 frames per second, which we aim at achieving further improvements in future work.
In addition, the bottleneck in localization may lie in the accuracy of 3D reconstruction and camera intrinsic parameters.
~\cref{fig:limitation} shows that the reconstruction from current SfM method~\cite{arandjelovic2016netvlad} can be noisy. 
More robust and efficient reconstruction methods will definitely benefit our system.

\vspace{-1mm}\section{Conclusion}\label{sec:conc}
\vspace{-2mm}
In this paper, we extend the scenarios of the camera re-localization problem to visually ambiguous extreme environments. 
Extensive experiments demonstrate that the ambiguity in such conditions brings great challenges to current visual localization methods.
To tackle the problem, we propose a hierarchical localization system by making use of temporal information and designing an environment-aware image enhancement framework.
Our system successfully localizes the ambiguous frames in those extreme scenes while retains competitive performance on the common indoor benchmark.

\clearpage

\bibliographystyle{ECCV/splncs04}
\bibliography{IEEEexample}
\end{document}